# AI-Generated Text Detection and Classification Based on BERT Deep Learning Algorithm

Hao Wang[1*], Jianwei Li[2], Zhengyu Li[3]

[1]Beijing Yuandian Technology Inc., Beijing, 102200, China
[2]Independent researcher, Beijing, 102200, China
[3]Nio Inc., Shanghai, 201805, China
*Corresponding author email: whdawn1006@gmail.com.

**Abstract.** With the rapid development and wide application of deep learning technology, AI-generated text detection plays an increasingly important role in various fields. In this study, we developed an efficient AI-generated text detection model based on the BERT algorithm, which provides new ideas and methods for solving related problems. In the data preprocessing stage, a series of steps were taken to process the text, including operations such as converting to lowercase, word splitting, removing stop words, stemming extraction, removing digits, and eliminating redundant spaces, to ensure data quality and accuracy. By dividing the dataset into a training set and a test set in the ratio of 60% and 40%, and observing the changes in the accuracy and loss values during the training process, we found that the model performed well during the training process. The accuracy increases steadily from the initial 94.78% to 99.72%, while the loss value decreases from 0.261 to 0.021 and converges gradually, which indicates that the BERT model is able to detect AI-generated text with high accuracy and the prediction results are gradually approaching the real classification results. Further analysis of the results of the training and test sets reveals that in terms of loss value, the average loss of the training set is 0.0565, while the average loss of the test set is 0.0917, showing a slightly higher loss value. As for the accuracy, the average accuracy of the training set reaches 98.1%, while the average accuracy of the test set is 97.71%, which is not much different from each other, indicating that the model has good generalisation ability. In conclusion, the AI-generated text detection model based on the BERT algorithm proposed in this study shows high accuracy and stability in experiments, providing an effective solution for related fields. In the future, the model performance can be further optimised and its potential for application in a wider range of fields can be explored to promote the development and application of AI technology in the field of text detection.

## 1. Introduction

AI-generated text detection refers to the use of AI technology to identify and detect text content generated by an AI system that may contain false information, misleading information, or content that violates regulations [1]. With the continuous development and popularisation of deep learning technology, AI-generated text detection plays an increasingly important role in the fields of network security, public opinion monitoring, news media, etc.

With the rapid development of AI technology, especially the breakthroughs in deep learning technology in recent years, various generative models such as GANs (Generative Adversarial

Networks) and RNNs (Recurrent Neural Networks) have been widely used in text generation tasks, which has made the problems of false information and fraudulent information increasingly serious [2]. The explosive growth of information on the Internet has resulted in a large amount of information that cannot be effectively managed and monitored, including a large amount of text content generated by AI systems [3]. Therefore, it is of great significance to carry out research on AI-generated text detection to maintain the order of cyberspace and protect the rights and interests of users. As social media platforms, news websites and other online platforms become the main channels for people to obtain information and exchange opinions, false information and rumours spread through AI-generated text also bring serious public opinion risks and social impacts [4].

Deep learning algorithms play a crucial role in AI-generated text detection. Deep learning algorithms excel in AI generated text detection tasks with their powerful pattern recognition and feature extraction capabilities [5]. Deep learning algorithms can capture the hidden patterns and features in text data by building complex neural network structures. For example, in the field of natural language processing, deep learning models such as the Transformer model have been widely used in text data processing tasks [6]. Deep learning algorithms can be trained on large-scale datasets to improve model performance. For AI-generated text detection tasks, having sufficiently diverse and authentic labelled data is a key factor in ensuring model accuracy and robustness. In addition, deep learning algorithms can be combined with word vector representations, attention mechanisms and other methods in natural language processing techniques to better understand and analyse text data and mine potential false information or offending content from it.

In the context of the current information explosion and the rapid development of artificial intelligence technology, it is of great significance to carry out research on AI-generated text detection to safeguard cyberspace security and prevent information fraud. And deep learning algorithms, as a powerful tool in this field, play an irreplaceable role in achieving efficient and accurate identification and filtering of false information. In this paper, an AI generated text detection model is developed based on BERT algorithm, which provides a new method for AI generated text detection.

## 2. Source of data sets

In this paper, experiments are conducted using a private dataset, which contains a large number of AI-generated texts as well as non-AI-generated texts, AI-generated texts labelled with 0 and non-AI-generated texts labelled with 1. There are 708 texts labelled with 0 and 670 non-AI-generated texts. The proportional distribution of the two is shown in Fig. 1, and the word counts of the manual and AI texts were statistically analysed using box-and-line plots, respectively, and the results are shown in Fig. 2. Part of the dataset is shown in Table 1.

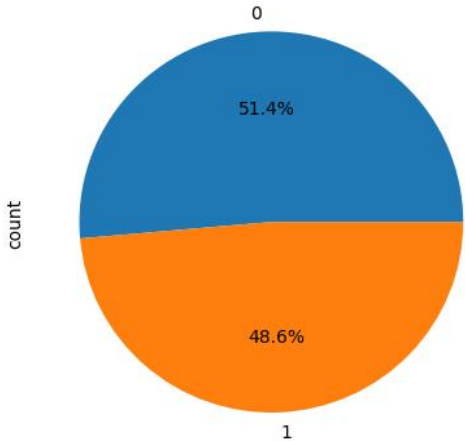

**Figure 1.** The proportional distribution of the two.
(Photo credit : Original)

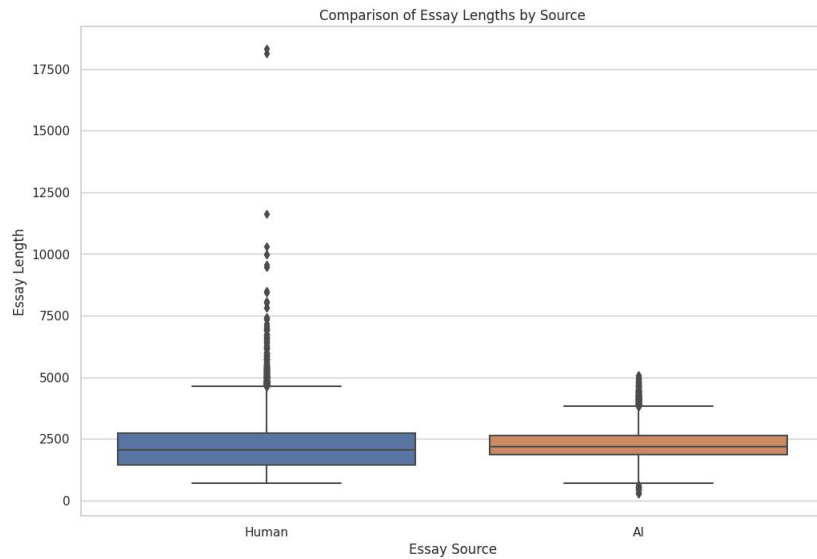

**Figure 2.** Box-and-line plots.
(Photo credit : Original)

**Table 1.** Part of the dataset.

| Text | Generated |
|---|---|
| Cars. Cars have been around since they became ... | 0 |
| Transportation is a large necessity in most co... | 0 |
| "America's love affair with it's vehicles seem... | 0 |
| How often do you ride in a car? Do you drive a... | 0 |
| Cars are a wonderful thing. They are perhaps o... | 0 |
| ... | ... |
| There has been a fuss about the Elector Colleg... | 0 |
| Limiting car usage has many advantages. Such a... | 0 |
| There's a new trend that has been developing f... | 0 |
| As we all know cars are a big part of our soci... | 0 |
| Cars have been around since the 1800's and hav... | 0 |

## 3. Text Preprocessing

Text preprocessing is a very important step in natural language processing that helps to clean and prepare text data for subsequent analysis or modelling. In this paper, the text is sequentially, converted to lowercase, word splitting, removal of stop words, stemming extraction, removal of numbers and removal of redundant spaces and other steps [7].

This paper firstly removes special characters, punctuation marks and other interfering information from the text to make the text cleaner. Secondly, all the text is converted to lower case form, which can avoid the same word being regarded as different words due to different case. After that the text is segmented into individual words or tokens, which is the basis for further processing of text data. Discontinued words are words that appear frequently in the text but have no real meaning, such as "the", "is", and so on. Removing these deactivated words can improve the effect of subsequent analyses [8]. For the English text, this paper carries out stemming extraction or word form reduction to convert the words to their basic forms, in order to reduce the impact of word distortion on the analysis. Eliminating redundant spaces and numbers helps to unify the format and make the data more standardised.

## 4. Method

GBERT (Bidirectional Encoder Representations from Transformers) is a pre-trained language model based on the Transformer architecture, proposed by Google.The principle of the BERT model mainly consists of two phases: pre-training and fine-tuning, which learns text data representations through large-scale unsupervised textual data representation, which can achieve excellent performance in various natural language processing tasks [9]. The structure of BERT model is shown in Figure 3.

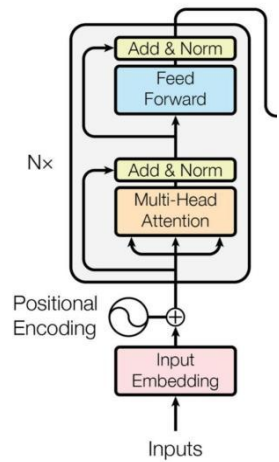

**Figure 3.** The structure of BERT model.
(Photo credit : Original)

Firstly, BERT implements a bidirectional encoder through the Transformer architecture, which can take into account the contextual information at the same time to better capture the relationship between words compared to the traditional unidirectional language model. This enables BERT to better understand the semantics and structure in a sentence when dealing with natural language tasks [10]. The BERT model utilises large-scale unlabelled text data for pre-training, in which two tasks, Masked Language Model (MLM) and Next Sentence Prediction (NSP), are used to learn the textual representation. The training process of the BERT model is shown in Figure 4.

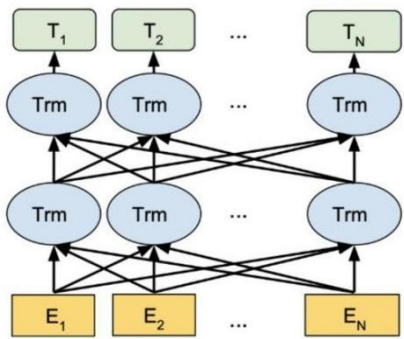

**Figure 4.** The training process of the BERT model.
(Photo credit : Original)

In the MLM task, a portion of the words in the input sequence is randomly masked out, and the model needs to predict these masked out words. This forces the model to model the contextual information and learn better word representations. And in the NSP task, the model needs to determine whether two sentences are consecutive or not. This task can help BERT learn the correlation between sentences to better understand the text sequence.

After pre-training, BERT can be fine-tuned for a variety of natural language processing tasks, such as text categorisation, named entity recognition, question and answer systems, etc. In the fine-tuning phase, one simply adds an output layer to the pre-trained model and combines it with labelled data for supervised learning to accomplish a specific task. Since BERT has learnt a common language representation, it only needs to fine-tune a small number of parameters to achieve better results when facing different tasks.

As a revolutionary natural language processing model, BERT has achieved state-of-the-art results in several benchmarks. Its power is not only reflected in its excellent performance, but also in its high flexibility and versatility. With the further development of BERT and its derivatives, we believe that it will play an increasingly important role in the field of natural language processing and bring us more surprises and breakthroughs.

## 5. Experiments and Results

The dataset is divided according to 60% and 40% and used as training set and test set respectively, the number of training rounds is set to 10, the graphic card used for the experiment is 3090, the GPU is 32G, and the experiment is conducted using python 3.10. The changes in accuracy and loss during the training process were recorded and the results are shown in Fig. 5.

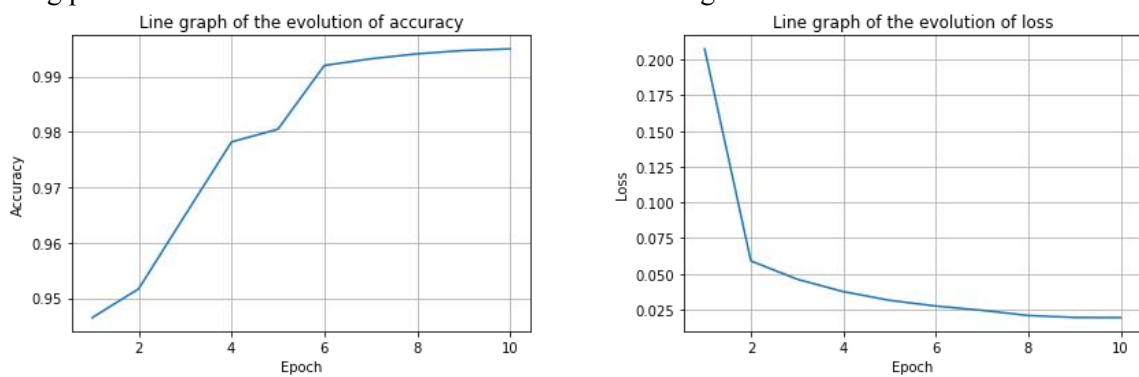

**Figure 5.** The changes in accuracy and loss during the training process.

(Photo credit : Original)

From the changes of accuracy and loss during the training process, it can be seen that the accuracy eventually grows from the initial 94.78% to 99.72%, which indicates that the BERT model can accurately detect AI-generated text, and the loss value decreases from the initial 0.261 to 0.021 and tends to converge, and the model's prediction is gradually approaching to the real classification results.

The average accuracy and average loss values of the training set and test set are output, and the results are shown in Table 2.

**Table 2.** The average accuracy and average loss.

|  | Loss | Accuracy |
| --- | --- | --- |
| Train | 0.0565 | 98.1 |
| Test | 0.0917 | 97.71 |

From the results of the training and test sets, the average loss of the training set is 0.0565 and the average loss of the test set is 0.0917, the test set has a higher loss compared to the training set. In terms of prediction accuracy, the average training accuracy of the training set is 98.1%, while the average accuracy of the test set is 97.71%, and the accuracy of the test set is decreased by 0.39% compared to the training set, which is not much different, indicating that the model has a good generalisation ability.

## 6. Conclusion

With the rapid development and wide application of deep learning technology, AI-generated text detection plays an increasingly important role in today's network security, public opinion monitoring, news media and other fields. In this study, an innovative AI-generated text detection model is constructed based on the BERT algorithm, which provides a novel solution for this field. In the data preprocessing stage, a series of steps were taken, including operations such as converting text to lowercase, performing word splitting, removing stop words, stemming extraction, and eliminating numbers and redundant spaces. By dividing the dataset into training and test sets in the ratio of 60% and 40%, we observe that the accuracy rate grows from the initial 94.78% to 99.72% during the training process, which indicates that the BERT model is able to detect the AI-generated text with high accuracy; at the same time, the loss value decreases from the initial 0.261 to 0.021 and tends to be stable, and the model's prediction results are gradually close to the real classification results.

Further analysis of the results of the training and test sets shows that the average loss of the training set is 0.0565, while the average loss of the test set is 0.0917, which shows that the test set has a slight upward trend relative to the training set. In terms of accuracy, the average accuracy of the training set reaches 98.1%, while the average accuracy of the test set is 97.71%, with only a 0.39% difference between the two. This indicates that the model has good generalisation ability and can achieve high accuracy on unknown data.

In conclusion, the AI-generated text detection model based on BERT algorithm proposed in this study performs well after sufficient training. By effectively processing and reasonably dividing the data and combining with excellent deep learning algorithms, it has achieved remarkable results in the field of AI-generated text detection. The model not only exhibits high accuracy and low loss value on the training set, but also shows relatively stable and excellent performance on the test set. This research result is not only of great significance for improving the level of AI-generated text detection technology, but also provides a useful reference for further research and practice in related fields. In the future, we can further expand the data scale, optimise the model architecture and explore more effective feature engineering methods, in order to further improve the effectiveness and reliability of AI-generated text detection technology in various application scenarios.